\documentclass{llncs}
\usepackage{amsfonts}
\usepackage{amsmath}
\usepackage{times}

\begin{document}

\title{Stability Analysis for Regularized Least Squares Regression}
\date{March 2003}
\author{Cynthia Rudin}
\institute{PACM, Fine Hall, Princeton University, Princeton NJ 08544, USA,\\
\email{crudin@princeton.edu},\\
\texttt{\small{http://www.math.princeton.edu/$^\sim$crudin}}}

\maketitle


\def\argminf{\hbox{\raise-1.6mm\hbox{$\textstyle
    \text{argmin} \atop f$\,}}}
\def\mthx{\mathbf{x}}
\def\X{\mathcal{X}}
\def\Y{\mathcal{Y}}
\def\L{\mathcal{L}}
\def\limt{\hbox{ \raise-1.0mm\hbox{$\textstyle
    \longrightarrow \atop ^{t \rightarrow \infty}$}}}
\def\limP{\hbox{ \raise1.0mm\hbox{$\textstyle
    _P \atop \longrightarrow$}}}
\def \sups{\hbox{\raise-2mm\hbox{$\textstyle
    sup \atop \|s\|_{H_{_1}}=1$}}}
\def \supc{\hbox{\raise-2mm\hbox{$\textstyle
    sup \atop \|c\|_{\ell_2}=1$}}}
\def \supf{\hbox{ \raise-2mm\hbox{$\textstyle
    sup \atop \|f\|_{\H}=1$}}}
\def\H{\mathcal{H}}
\def\F{\mathcal{F}}
\def\R{\mathbb{R}}
\def\E{\mathbb{E}}
\def\Z{\mathcal{Z}}
\def\sumi{\sum_{i=1}^N}
\def\sumit{\sum_{i=1}^{N_t}}
\def\suml{\sum_{\ell=1}^N}
\def\sumj{\sum_{j=1}^N}
\def\sumij{\sum_{i,j=1}^N}
\def\summn{\sum_{m,n=1}^N}
\def\regterm{\lambda\|f\|_\H^2}
\def\minimizef{\hbox{\raise-1.6mm\hbox{$\textstyle
    \text{minimize} \atop f$}}}
\def\maxj{\hbox{\raise-1.6mm\hbox{$\textstyle
    \text{max} \atop j$}}}

\begin{center}{March 7, 2003}\end{center}

\begin{abstract}
We discuss stability for a class of learning algorithms with
respect to noisy labels. The algorithms we consider are for
regression, and they involve the minimization of regularized risk
functionals, such as $L(f) := \frac{1}{N} \sumi
(f(\mthx_i)-y_i)^2+\regterm$. We shall call the algorithm `stable'
if, when $y_i$ is a noisy version of $\overline{f}(\mthx_i)$ for
some function $\overline{f} \in \H$, the output of the algorithm
converges to $\overline{f}$ as the regularization term and noise
simultaneously vanish. We consider two flavors of this problem,
one where a data set of $N$ points remains fixed, and the other
where $N \rightarrow \infty$. For the case where $N \rightarrow
\infty$, we give conditions for convergence to $f_{\E}$ (the
function which is the expectation of $y(\mthx)$ for each $\mthx$),
as $\lambda\rightarrow 0$. For the fixed $N$ case, we describe the
limiting 'non-noisy', 'non-regularized' function $\overline{f}$,
and give conditions for convergence. In the process, we develop a
set of tools for dealing with functionals such as $L(f)$, which
are applicable to many
other problems in learning theory. 
\end{abstract}

\small{\noindent \textbf{keywords} statistical learning theory,
learning in the limit, regularized least squares regression,
RKHS}\\

\normalsize
\section{Introduction}

In regression learning problems, we are given data
$(\mthx_i,y_i)_{i=1,..,N}$ in $\X \times \Y$ where $\X$ is a
bounded subset of $\R^n$ and $\Y$ is a bounded subset of $\R$. We
assume this data is chosen iid (independently and identically
distributed) according to an unknown probability distribution
$\mu(\mthx,y)$. We say that $\mthx$ is a `position', and $y$ is a
`label'. These data points may be, for example, images of people's
faces in pixel space with a person's age as the corresponding
label, or auto-regressive time series data (\cite{Poggio-02},
\cite{Sayan-timeseries97}). The output of a learning algorithm is
a decision function $f: \X \rightarrow \R$. Even though we only
know $N$ data points from distribution $\mu(\mthx,y)$, we hope to
construct $f$ which will be able to \textit{generalize} to
unobserved points in the distribution. This means we would like
$f$ to predict the value of $y$ for any given value of $\mthx \in
\X$. Since we want our function $f$ to fit the data accurately and
also have this generalization ability, we refer to Vapnik's
Structural Risk Minimization (SRM) principle (\cite{VapSLT},
\cite{VapNSLT}). In SRM, we limit our choice of functions $f$ so
they are chosen from a class $\F$, of finite 'capacity' (i.e.
finite VC dimension). Otherwise, we cannot hope to choose a
function $f$ which has generalization ability - we would overfit
the data. One convenient way to implement SRM is to let $\F$ be a
ball within a Reproducing Kernel Hilbert Space (RKHS) $\H$, with
norm $\|\cdot\|_\H$. In this form, we have an Ivanov
regularization problem; one can show that the solution is always
the minimizer of a corresponding Tikhonov regularization problem.
Algorithms for classification and regression solve this Tikhonov
Regularization problem, so that the decision function is given by
$f_{min}$ (\cite{Smale-02}, \cite{Sch}), where
\begin{equation*}
f_{min} := \argminf L(f)\text{, where } L(f)= \frac{1}{N}\sumi
V(f(\mthx_i)-y_i)+\regterm \;.
\end{equation*}
\noindent $L(f)$ is called the Regularized Risk functional. Note
that we define our RKHS $\H$ so that $f \in \H$ iff $\|f\|_\H$ is
finite. Thus, minimizing over $f \in \H$ is equivalent to
minimizing over all functions $f$. The first term in $L(f)$ is
called the Empirical Risk, and $V( \cdot )$ is a pre-determined
loss function. We will generally use the least squares loss
function $V(z)=(z)^2$, but a similar analysis can be performed for
other loss functions. The second term is called the Regularization
term, and $\lambda$ is called the Regularization parameter; one
always takes $\lambda>0$. Here, $\lambda$ can be viewed as the
trade-off between accuracy and generalization. If $\lambda$ is
very small, we are minimizing the Empirical Risk, increasing the
accuracy of our model to the data, and possibly overfitting. If
$\lambda$ is very large, our algorithm will generalize at the
expense of accuracy.
In a sense, $\lambda$ controls the capacity of the function class
from which $f$ is chosen: the larger $\lambda$ is, the smaller the
radius of the ball $\F$ in $\H$. In practice, $\lambda$ is often
chosen empirically, perhaps to minimize the leave-one-out error on
a training set.

Another interpretation of this functional is through the eyes of
algorithmic stability, as described by Bousquet and Elisseeff
(\cite{BE-01}). Here, the regularization term prevents the
algorithm from being sensitive to the replacement of one data
point. In either case, the regularization problem is well-posed
only when $\lambda$ is strictly greater than 0.

We assume that the labels $y$ are `noisy', in the sense that there
is a marginal distribution $\mu(y|\mthx)$ for each $\mthx$. We
denote the expectation value of the label y for position x as
$\E(y|\mthx)$, and we denote the marginal distribution along the
$\mthx$-axis as $\mu(\mthx)$. (This is the distribution of
$\mu(\mthx,y)$ after integrating over the $y$ values.)

For the case when $N\rightarrow \infty$, we show convergence of
$f_{min}$ to a function $f_{\E}$ as the regularization term
vanishes, provided $f_{\E} \in \H$; i.e. we need to find
conditions on the simultaneous convergence $N\rightarrow \infty$
and $\lambda \rightarrow 0$ so that $f_{min} \rightarrow f_{\E}$.
Here, the function $f_{\E}$ is defined by:
\begin{eqnarray*}
f_\E  = \argminf \textit{Actual Risk}\,(f) \text{, where }
\textit{Actual Risk}\,(f) =  \int
\left(\E[f(\mthx)-y]^2|\mthx\right) d\mu(\mthx).
\end{eqnarray*}
\noindent In other words, $f_\E$ is the minimizer of the
\textit{`Actual Risk'}. Since we are using the least squares loss
function, this minimizer is simply the expectation of $y$ for each
$\mthx$; $f_\E(\mthx)= \E(y|\mthx)$.

We assume that we have chosen a RKHS which is large enough to
contain $f_\E(\mthx)$. In other words, $\|f_\E(\mthx)\|_\H
<\infty$. This is not an exceedingly strong assumption; in fact,
many popular kernels (e.g. gaussian kernels) can produce RKHS of
arbitrarily high VC dimension. Although $f_\E(\mthx)$ may not be
in $\H$ for every case, $f_\E(\mthx)$ will be in $\H$ for most
smooth processes which have bounded noise, as long as we implement
a sufficiently powerful RKHS.

For the fixed $N$ case, we may express label $y$ for position
$\mthx$ as the random variable $y(\mthx) = \tilde{f}(\mthx) +
b(\mthx)$, where $\tilde{f}:\X \rightarrow \Y$ is a deterministic
function assumed to be in $\H$, and $b(\mthx)$ is random noise
with some probability distribution, with $b(\mthx)$ and
$b(\mthx')$ independent if $\mthx \neq \mthx'$. We denote the
vector of noise values as $\mathbf{b} = (b_1, b_2,..,b_N) =
\{b(\mthx_i)\}_{i=1,..,N}$. In order to force the noise to vanish,
we will assume the noise is generated by a fixed random process
generating noise with norm bounded by $b_{max}$ almost surely, and
we will only shrink its amplitude. Since the noise is generated by
this fixed process, the theorem will hold whenever the noise is
bounded, and thus, if the noise is bounded almost surely, the
theorem will hold almost surely. Using the least squared loss and
making the noise explicit, our algorithm becomes:
\begin{eqnarray*}
f_{min} := \argminf L(f) \text{, where } L(f) &=&
\frac{1}{N}\sumi(f(\mthx_i)-\tilde{f}(\mthx_i)-c\:
b_i)^2 + \lambda \|f\|_H^2,\\
\text{where }\|\mathbf{b}\|_{\ell_2} &\leq& \sqrt{N} b_{max}
\text{ and } c \text{ is a constant.}
\end{eqnarray*}
For $\lambda>0$, the minimizer of $L(f)$ is unique, because
$\lambda \|f\|^2_\H$ is strictly convex. Since the noise is
random, $f_{min}$ is still a random variable. Our goal is to show
`stability' for this algorithm, i.e. we need to find a set of
conditions on the simultaneous convergence $\lambda, c \rightarrow
0$ which allows $f_{min} \rightarrow \overline{f}$, where
$\overline{f}$ is the element of $\H$ with minimal norm that has
zero Empirical Risk when noise is not present. (Since we assume
that $\tilde{f} \in \H$, $\tilde{f}$ itself minimizes $L(f)$ when
$\lambda=0$. Since many functions in $\H$ vanish at all the
$\mthx_i$, there may be infinitely many functions with zero
empirical risk; our algorithm will converge to the one with the
smallest RKHS norm.)

Intuitively, this stability analysis demonstrates that there's no
inherent error in our algorithm when noise or regularization is
present, and that a small amount of noise or regularization cannot
dramatically disrupt the algorithm's output. This type of
stability is different from the `algorithmic stability' of
Devroye(\cite{Devroye}). Algorithmic stability measures the
variability of an algorithm's output as the data set changes. Our
type of stability determines whether the algorithm's output
changes dramatically when noise or regularization is present.
Algorithmic stability is a property of one particular algorithm
for one particular distribution of data. Our stability is not -
the distribution changes as noise is removed, and the algorithm
changes as the regularization term shrinks. We actually use
algorithmic stability to help us show stability of our algorithm
in this sense.

Theorem 1 states that the regularized least squares regression
algorithm is stable as the number of data points increases to
infinity. Theorem 2 states that the regularized least squares
regression algorithm is stable for a fixed N point data set.

\noindent \hrulefill\\
\noindent\textbf{Main Algorithm} (Regularized Least Squares
Regression): \\
For a data set $Z=(\mthx_i,y_i)_{i=1,..,N}$, where $\forall i\in
1,..,N$, $\mthx_i \in \R^n$, $y_i \in \R$
\begin{eqnarray*}
f_{Z,\lambda} := \argminf L_{Z,\lambda}(f)\text{, where
}L_{Z,\lambda}(f) &=& \frac{1}{N}\sumi(f(\mthx_i)-y_i)^2 + \lambda
\|f\|_\H^2\;.
\end{eqnarray*}
\noindent \hrulefill
\begin{theorem}
Denote by $f_\E(\mthx)$ the function $\E(y|\mthx)$. Denote by
$Z_N$ the data set $(\mthx_i,y_i)_{i=1,..,N}$. If $f_\E\in\H$ and
if $\lambda:=\lambda_N$ is chosen to depend on $N$ such that
$\lambda_N \rightarrow 0$ and $N\lambda_N^3\rightarrow\infty$ as
$N \rightarrow \infty$, then we have convergence of the
\textit{Main Algorithm}:
\begin{equation*}
\|f_{Z_N,\lambda_N}-f_\E\|_\H \limP 0 \text{ as } N \rightarrow
\infty.
\end{equation*}

\noindent Here, `$\limP$' denotes convergence in probability.
\end{theorem}
\noindent\hrulefill

\begin{theorem}
Assume we are given $N$ fixed positions $\mthx_1,..,\mthx_N$.
\item Suppose that for each $i\in 1,..,N$, the labels are given by
$y_i = \tilde{f}(\mthx_i) + \frac{1}{t} b_i$, where the $b_i$'s
are independent random variables with
$\|\mathbf{b}\|_{\ell_{_2}}\leq \sqrt{N} b_{max}$ almost surely.
Denote by $Z_t$ the data set $(\mthx_i,y_i)_{i=1,..,N}$.

Define $\overline{f}$ by:\\
\noindent \vspace{0.1pt} \hspace{20pt} \textbf{(i)}
$\overline{f}(\mthx_i)=\tilde{f}(\mthx_i)$ for $i=1,..,N$\\
\noindent \vspace{0.1pt}\hspace{20pt} \textbf{(ii)}
$\|\overline{f}\|_\H$ is minimal, among all functions which
satisfy \textbf{(i)}.\\

If $\lambda:=\lambda_t$ is chosen to depend on $t$ such that
$t\sqrt{\lambda_t} \rightarrow \infty$ as $t \rightarrow \infty$
and $\lambda_t \rightarrow 0$, then we have convergence of the
\textit{Main Algorithm} almost surely:
\begin{equation*}
\|f_{Z_t,\lambda_t}-\overline{f}\|_\H \rightarrow 0 \text{ as } t
\rightarrow \infty \text{\;\;\;\;almost surely .}
\end{equation*}
\end{theorem}
\noindent \hrulefill\\

Section 2 contains a short review of RKHS. Section 3 and 4 contain
the proofs of Theorems 1 and 2.

\section{Reproducing Kernel Hilbert Space (RKHS)}

$\H$ is a real Reproducing Kernel Hilbert Space (RKHS) if $\H$ has
the following properties:
\begin{itemize}
\item $\H$ \textit{is a Hilbert space. } $\H$ is a complete, inner
product, real vector space of functions $f:\X\rightarrow\R$. We
denote $\H$'s inner product by $(\cdot,\cdot)_\H$, and $\H$'s norm
by $\|\cdot\|_\H$.

\item \textit{Reproducing Property. } There exists a bilinear form
$K:\X\times\X \rightarrow \R$ \;such that $\forall \mthx \in \X$,
we have $K(\mthx,\cdot)\in \H$ and $(f,K(\mthx,\cdot))_\H =
f(\mthx)$ for any $f \in \F$. This $K$ is called the `reproducing
kernel' of the RKHS.\cite{Sch}\cite{Tikhonov}\cite{Smale-02}. We
sometimes denote $K(\mthx,\cdot)$ by $K_\mthx$.

\item \textit{Spanning Property. } $\H =
\overline{span\{K(\mthx,\cdot)| \mthx \in \X\}}$
\end{itemize}

\noindent Since $\H$ is a real Hilbert space, $(f,g)=(g,f)$ for
all $f,g\in\H$. It follows that
$K(\mthx,\mthx')=(K(\mthx,\cdot),K(\mthx',\cdot))_\H =
(K(\mthx',\cdot),K(\mthx,\cdot))_\H=K(\mthx',\mthx)$, i.e. $K$ is
symmetric in its two arguments. An equivalent definition of an
RKHS is a Hilbert space of functions $f: \X \rightarrow \R$ such
that all evaluation functionals $\Gamma_\mthx : f \rightarrow
f(\mthx), \;\mthx\in\X$, are continuous. Given $\mthx_1,..,\mthx_N
\in \X$, the associated $N \times N$ Gram Matrix $G$ has entries
$G_{ij} = K(x_i,x_j)$ where $K$ is the reproducing kernel for the
RKHS $\H$. The Gram Matrix is always a positive semi-definite
matrix.

The Representer Theorem transforms the minimization of our
functional $L_{Z,\lambda}(f)$ into an optimization problem over
only $N$ numbers. This advantage is the main reason why scientists
take $\F$ to be a ball in a RKHS $\H$. We present a corollary of
this theorem below. \\

\noindent\textbf{Corollary of the Representer Theorem} (Kimeldorf,
Wahba)\cite{Wahba-68})
   \textit{The function $f_{min}=argmin_f \;\frac{1}{N} \sumi V(f(\mthx)-y_i) + \regterm$ can be
   represented in the form $f_{min}= \sumi \alpha_i K_{x_i}$. This is true for any
   arbitrary loss function V.}\\
(This corollary is a specific case of the full Representer Theorem
\cite{Wahba-68}.)\\

Having described the basic facts about RKHS, we now continue with
the proofs of Theorems 1 and 2.

\section{Proof of Theorem 1}

\indent For the \textit{Main Algorithm} above, we are increasing
the size of the data set $Z_N$ as $N$ increases. We need to show
convergence $f_{Z_N,\lambda_N} \limP f_\E$, where $\lambda_N
\rightarrow 0$ and $N \rightarrow \infty$. That is, we need to
show
\begin{equation*}
\lim_{N\rightarrow\infty} P\{\|f_{Z_N,\lambda_N}-f_\E\|_\H \geq
\eta\} = 0 \text{ for every }\eta>0\;.
\end{equation*}
We can break up the distance $\|f_{Z_N,\lambda_N}-f_\E\|_\H$ into
two contributions. The first contribution is called `variance',
and it is due to the finite number of randomly chosen noisy data
points. The variance vanishes with arbitrarily high probability as
the number of data points increases, even if the noise does not
vanish. The second contribution is the `bias' due to the
restriction we place on our hypothesis space, i.e. the fact that
$f$ is chosen from with a ball of a RKHS. This term vanishes as
the ball gets larger, i.e. when $\lambda_N$ gets smaller.

\begin{eqnarray*}
\begin{array}{ccccc}
  \|f_{Z_N,\lambda_N}-f_\E\|_\H & \leq &
  \underbrace{\|f_{Z_N,\lambda_N}-\hat{f}_{\lambda_N}\|_\H} &+&
  \underbrace{\|\hat{f}_{\lambda_N} - f_\E\|_\H}\\
  &&\text{variance} && \text{bias} \\
\end{array}
\end{eqnarray*}
\noindent where $f_{Z_N,\lambda_N} = \argminf
L_{Z_N,\lambda_N}(f),$
\begin{center}where $L_{Z_N,\lambda_N}(f) = \frac{1}{N}\sumi
(f(\mthx_i)-y_i)^2 + \lambda_N \|f\|_H^2$, and\end{center}
$\hat{f}_{\lambda_N} = \argminf \hat{L}_{\lambda_N}(f)$, where
$\hat{L}_{\lambda_N}(f) = \int (\E[f(\mthx)-y]^2|\mthx)d\mu(\mthx)
+ \lambda_N \|f\|_H^2.$

Lemma 1.1 below describes a method for proving that
the minimizers of two convex functions are close in $\H$.\\

\noindent\textbf{Lemma 1.1.} \textit{Suppose $L^1, L^2:\H
\rightarrow \R$ are two convex functionals for which there exist
$\varepsilon$,$\delta$ so that:
\begin{description}
    \item[(a)] $\forall\; f \in \H \;;\;|L^1(f) - L^2(f)| <
    \varepsilon$
    \item[(b)] $|L^1(f) - L^1(f^1)| < 2 \varepsilon  \Longrightarrow
    \|f-f^1\|_\H <\delta$
\end{description}
Then if the minimizers $f^1 := \argminf L^1(f)$ and $f^2 :=
\argminf L^2(f)$ exist, they satisfy $\|f^1-f^2\|_\H <\delta$.}

\begin{proof}
Since $f^1$ and $f^2$ are minimizers of $L^1$ and $L^2$
respectively, and using the closeness condition \textbf{(a)}:
\begin{center}
$L^1(f^1) \leq L^1(f^2) \leq L^2(f^2) +\varepsilon$\\
$L^2(f^2) \leq L^2(f^1) \leq L^1(f^1) + \varepsilon$\\
So, $|L^1(f^1) - L^2(f^2)| \leq \varepsilon.$
\end{center}
Now, $|L^1(f^1)-L^1(f^2)| \leq
|L^1(f^1)-L^2(f^2)|+|L^2(f^2)-L^1(f^2)| \leq 2\varepsilon$, and
finally, condition \textbf{(b)} will give us $\|f^1-f^2\|_\H
\leq\delta$. \qed
\end{proof}

\noindent Back to the proof of Theorem 1. We proceed one term at a time.\\

\noindent \textbf{Variance Term} We will choose more general
versions of $L_{Z_N,\lambda_N}(f)$ and $\hat{L}_{\lambda_N}(f)$
temporarily.
\begin{eqnarray*}
\L_{Z_N,\lambda_N}(f) &:=& \frac{1}{N} \sumi
V(f(\mthx_i)-y_i) + \lambda_N \|f\|_\H^2\\
\hat{\L}_{\lambda_N}(f)&:=& \int \E (V(f(\mthx)-y)|\mthx)
d\mu(\mthx) + \lambda_N \|f\|_\H^2
\\&\text{where}&\; |V(a) - V(b)| \leq \sigma_V |a-b|.
\end{eqnarray*}
That is, we assume that the loss function $V$ is Lipschitz
continuous, or `sigma-admissible' \cite{BE-01}. The least squares
loss has $\sigma_V = 2 \X_{max}$, since $|V(a)-V(b)| = |a^2-b^2|
\leq |a+b||a-b| \leq 2 \X_{max} |a-b|$.

We need to verify the conditions \textbf{(a)} and \textbf{(b)} in
order to use Lemma 1.1. To verify the closeness property
\textbf{(a)} for our functionals $\L_{Z_N,\lambda_N}(f)$ and
$\hat{\L}_{\lambda_N}(f)$:
\begin{eqnarray}\nonumber
\Big|  \L_{Z_N,\lambda_N}(f) - \hat{\L}_{\lambda_N}(f)\Big|&=&
\Big|\frac{1}{N} \sumi V(f(\mthx_i)-y_i) - \int \E
(V(f(\mthx)-y)|\mthx) d\mu(\mthx)\Big|\\ &=&
\Big|\;\;\;\textit{Empirical Risk } (f) \;\;-\;\; \textit{Actual
Risk } (f)\;\;\;\Big| \label{RempRactual}
\end{eqnarray}

\noindent There are many available upper bounds for the right side
of equation (\ref{RempRactual}), including Vapnik's VC bound,
which relies on the VC dimension of the class of allowed decision
functions $\F$ (\cite{VapSLT}, \cite{VapNSLT}). The particular
bound we utilize for this paper was constructed by Bousquet and
Elisseeff \cite{BE-01}, and it is based on `algorithmic
stability'. In general, bounds of this quantity are probabilistic,
and are based on some capacity measure of the algorithm or space
of functions $\F$. This particular bound relies on the
sigma-admissibility of the loss function V, and McDiarmid's
concentration of measure inequality.\\

\noindent \underline{def}\;\;\; $\Z_{N}$ is a training sample
${(\mthx_1,y_1),..,(\mthx_{N},y_{N})}$.
$\Z_{N;\underline{\mthx},\underline{y}}^i =
(\Z_{N}\backslash{(\mthx_i,y_i)})\cup
{(\underline{\mthx},\underline{y})}$. That is, we replace the
$i^\text{th}$ training point in $\Z_{N}$ by a new
data point in order to obtain $\Z_{N;\underline{\mthx},\underline{y}}^i$.\\


\noindent \underline{def}\;\;\; The algorithm $Alg : \Z
\rightarrow f_\Z$ is \textit{uniformly $\beta$-stable with respect
to loss function} $V_\beta : \X \times \Y \times \R \rightarrow
\R$ if: $|V_\beta(\mthx,y,f_{\Z_{N}}(\mthx)) -
V_\beta(\mthx,y,f_{\Z_{N;\underline{\mthx},\underline{y}}^i}(\mthx))|
\leq \beta \text{ for all } (\mthx,y), \;
(\underline{\mthx},\underline{y}) \in \X \times \Y, i, \text{ and
all } \Z_{N}.$\\

\noindent Basically, this algorithmic stability measures how much
the algorithm's output could possibly change, as measured by the
loss function $V_\beta$, when we replace one data point.\\

\noindent\textbf{Algorithmic Stability Theorem} (Bousquet and
Elisseeff, \cite{BE-01}) \textit{If we are given a uniformly
$\beta$-stable algorithm with respect to loss function $V_\beta$,
which outputs functions bounded by the constant $M$ (i.e.
$|f_Z(\mthx)|\leq M \;\;\forall\; \mthx \in \X,\; \forall\Z$),
then for any $N\geq\frac{8M^2}{\varepsilon^2}$,}
\begin{eqnarray*}
P\{|\textit{Empirical Risk}(f)-\textit{Actual
Risk}(f)|>\varepsilon\}\leq p_N, \text{with }p_N=\frac{64 M N
\beta + 8 M^2}{N \varepsilon^2}.
\end{eqnarray*}

\noindent\textbf{Algorithmic Stability of Tikhonov Learning
Algorithms} (Bousquet and Elisseeff, \cite{BE-01}) \textit{The
\textit{Main Algorithm} is uniformly $\beta$-stable, with
$\beta=\frac{C^2 \kappa^2}{N \lambda_N}$. Here, $C$ is an upper
bound on the sigma-admissibility constant $\sigma_{V}$, and
$\kappa$ is an upper bound on the diagonal
elements of the Gram Matrix, that is, $max_i \;\;G_{ii} \leq \kappa$.}\\

Returning to the proof of Theorem 1, we now know the right side of
(\ref{RempRactual}) is bounded by $\varepsilon$ (for large values
of $N$) with probability at least $1-p_N$, where:
\begin{eqnarray}\nonumber
p_N=\frac{64 M N \beta + 8 M^2}{N \varepsilon^2}\text{, where }\:
\beta = \frac{C^2 \kappa^2}{N \lambda_N} \text{, so } p_N =
\frac{64 M C^2 \kappa^2 + 8 M^2 \lambda_N}{N \lambda_N
\varepsilon^2}. \label{thm1p}
\end{eqnarray}

We now have the closeness condition \textbf{(a)} of Lemma 1.1
satisfied with probability at least $1-p_N$, i.e. $|
L_{Z_N,\lambda_N}(f) - \hat{L}_{\lambda_N}(f) | \leq \varepsilon$,
with probability at least $1-p_N$, where $p_N$ is given in
(\ref{thm1p}).

Now we verify condition \textbf{(b)}. We need to show that
$|\hat{L}_{\lambda_N}(f)-\hat{L}_{\lambda_N}
(\hat{f}_{\lambda_N})| \leq 2\varepsilon$ implies that
$\|f-\hat{f}_{\lambda_N}\|_\H \leq \delta$ for every $f$. Let's
define a function $h$ so that $h := f - \hat{f}_{\lambda_N}$ in
what follows.
\begin{eqnarray*}
\hat{L}_{\lambda_N}(f) &=& \int\E(\hat{f}_{\lambda_N}(\mthx)+
h(\mthx)-y\,|\,\mthx)^2d\mu(\mthx)
+ \lambda_N\|\hat{f}_{\lambda_N} + h\|_\H^2\\
 &=& \hat{L}_{\lambda_N}(\hat{f}_{\lambda_N}) + \Big[ 2
\int\E [(\hat{f}_{\lambda_N}(\mthx)-y)h(\mthx)|\mthx]d\mu(\mthx)
+\; 2\lambda_N (\hat{f}_{\lambda_N},h)_\H \Big] \\&& + \int
 h^2(\mthx) d \mu(\mthx)+ \lambda_N\|h\|_\H^2
\end{eqnarray*}

\noindent The terms in the brackets are linear in $h$. Remember
that $\hat{f}_{\lambda_N}$ is the minimizer of
$\hat{L}_{\lambda_N}$, and thus the linear terms must be zero. (If
the linear terms are non-zero, we can reverse the sign of $h$ and
contradict $\hat{f}_{\lambda_N}$ as the minimizer.) The last two
terms are always positive.
\begin{eqnarray*}
\hat{L}_{\lambda_N}(f) &=&
\hat{L}_{\lambda_N}(\hat{f}_{\lambda_N})+ \int h^2(\mthx) d
\mu(\mthx) + \lambda_N\|h\|_\H^2 \geq
\hat{L}_{\lambda_N}(\hat{f}_{\lambda_N}) + \lambda_N\|h\|_\H^2
\end{eqnarray*}
Now we can see that \textbf{(b)} holds:
\begin{eqnarray*}
\|f-\hat{f}_{\lambda_N}\|_\H^2 = \|h\|_\H^2 \leq
(\hat{L}_{\lambda_N}(f)-\hat{L}_{\lambda_N}
(\hat{f}_{\lambda_N}))\frac{1}{\lambda_N} \leq \frac{2
\varepsilon}{\lambda_N}
\end{eqnarray*}
In this case,
\begin{equation} \label{thm1variancebd}
\delta = \sqrt{\frac{2 \varepsilon}{\lambda_N}}\;.
\end{equation}

\noindent Since both the conditions \textbf{(a)} and \textbf{(b)}
are satisfied, Lemma 1.1 produces
\begin{eqnarray*}
\|f_{Z_N,\lambda_N} - \hat{f}_{\lambda_N}\|_\H \leq
\sqrt{\frac{2\varepsilon}{\lambda_N}} \text{ with prob. at least }
1-p_N,
\end{eqnarray*}
where $p_N$ is given in (\ref{thm1p}). We are done with the variance term.\\

\noindent \textbf{Bias Term} We will prove that the bias term
vanishes using the spectral theorem. Define the function
$h(\mthx):=f(\mthx)-f_\E(\mthx)$ in what follows. Now,
\begin{eqnarray*}
\hat{L}_{\lambda_N}(f)
&=& \int (K_\mthx,h)_\H^2 d\mu(\mthx) +\int \E[(f_\E(\mthx) -
y)^2|\mthx]d\mu(\mthx)+ \lambda_N \|h +f_\E\|^2_\H
\end{eqnarray*}

The minimizer of $\hat{L}_{\lambda_N}(f)$ again must have first
variational derivative equal to 0. Using Fubini's Theorem, we
find:

\begin{eqnarray*}\frac{\partial \hat{L}_{\lambda_N} (f+\gamma g)}
{\partial \gamma}\bigg|_{\gamma=0} &=&\frac{\partial}{\partial
\gamma} \left[ \int (K_\mthx,h+\gamma g)_\H^2 d \mu(\mthx) +
\lambda_N \|h+f_\E + \gamma g\|_\H^2\right]\bigg|_{\gamma=0}\\
&=& 2\left(\int (K_\mthx,h)K_\mthx d \mu(\mthx) + \lambda_N
(h+f_\E),g\right)_\H
\end{eqnarray*}

\noindent If $f=\hat{f}_{\lambda_N}$, the above expression must be
zero for all $g$, thus:
\begin{eqnarray}\label{thm1T}
0=\int (K_\mthx,\hat{f}_{\lambda_N}-f_\E)_\H K_\mthx d \mu(\mthx)
+ \lambda_N (\hat{f}_{\lambda_N})
\end{eqnarray}

\noindent Let's define a new operator $T$.\\
\noindent $\text{\underline{def}}\;\; T: \;\;\H \;\longrightarrow \H \\
\text{}\hspace{45pt}f \longmapsto \int (K_\mthx,f)_\H K_\mthx d
\mu(\mthx)$\\

\noindent One can check that $T$ is self-adjoint since
$(Tf,g)_\H=\int(K_\mthx,f)_\H(K_\mthx,g)_\H
d\mu(\mthx)=(f,Tg)_\H$. For an operator $Q$ from one Hilbert space
$H_1$, to another, $H_2$, the operator norm of $Q$ is defined by
$\|Q\|_{\mathcal{L}(H_1,H_2)} := \sups \|Q s\|_{H_2}$. Our
operator $T$ is bounded, since by Cauchy-Schwarz,
\begin{eqnarray*}
\|T\|_{\mathcal{L}(\H,\H)}^2
&\leq&\supf\|f\|_\H^2\int\int \sqrt{K(\mthx,\mthx)}
K(\mthx,\mathbf{z})\sqrt{K(\mathbf{z},\mathbf{z})}d
\mu(\mthx)d\mu (\mathbf{z}) \; \leq \; \infty.\\
\end{eqnarray*}

We are going to use the spectral theorem next, but first let us
review a few facts from functional analysis about this theorem
(\cite{RS}). The spectral theorem allows one to define functions
of a bounded self-adjoint operator on a Hilbert space $H$. If the
function is a polynomial, e.g. $f(z) = 3x^2-5z+2$, then it is
clear how to define the corresponding operator $\phi(A):
\phi(A)=3A^2 - 5A + 2$. The spectral theorem extends the
correspondence $\phi(z) \leftrightarrow \phi(A)$ to all continuous
functions (in fact, to all bounded Borel functions). Moreover, one
has $\|\phi(A)\|_{\mathcal{L}(H,H)} \leq sup \{ |\phi(z)|\; ; \; z
\in spec(A)\}$. Because $\phi$ is a real function, the operator
$\phi(A)$ provided by the spectral theorem is also self-adjoint.
In addition, for each $f \in H$, we have a measure $\nu_{f;A}$ on
$spec(A)$ such that $(\phi(A)f,f)_H = \int_{spec(A)} \phi(z) d
\nu_{f;A}(z)$. The measure $\nu_{f;A}$ is concentrated on that
part of the spectrum $spec(A)$ along which $f$ has a nonzero
component. In particular, if $f \in Ker(A)$, then $f$ is an
eigenvector of $A$ with eigenvalue 0, and $\nu_{f;A}$ is a
$\delta$-measure concentrated on $\{0\}$. If on the contrary, $f
\perp Ker(A)$, then $\nu_{f;A}(\{0\})=0$.

Now, using the definition of the spectral measure $\nu_{f_\E;T}$
for the operator $T$ and the function $f_\E$, we have from
(\ref{thm1T}):
\begin{eqnarray*}
0 &=& T(\hat{f}_{\lambda_N}-f_\E) + \lambda_N (\hat{f}_{\lambda_N})\\
\Rightarrow \;\;\hat{f}_{\lambda_N}-f_\E &=&  (T+\lambda_N)^{-1} (-\lambda_N f_\E) \\
\Rightarrow \;\;\|\hat{f}_{\lambda_N}-f_\E\|_\H^2 &=&
\|(T+\lambda_N)^{-1}\lambda_N f_\E\|_\H^2 = \int_{spec(T)} \left(
\frac{\lambda_N}{\gamma+\lambda_N} \right)^2 d
\nu_{f_\E;T}(\gamma)
\end{eqnarray*}
Since $K(\cdot,\cdot)$ is positive semidefinite, $T$ is a positive
operator and thus has non-negative spectrum only. One can see that
$Ker\;T$ is empty, i.e. take any function $\vartheta$ such that $T
\vartheta=0$. Then, $0 = (T \vartheta,\vartheta)_\H = \int
\vartheta^2(\mthx)d \mu(\mthx)$; thus, $\vartheta$ must be zero
almost everywhere. It follows that $\{0\}$ is a set of measure
zero for $\nu_{f_\E;T}$. As $N\rightarrow\infty$, $\lambda_N
\rightarrow 0$, and the function
$(\frac{\lambda_N}{\gamma+\lambda_N})$ converges to 0 pointwise on
$\R_+$, and thus almost everywhere with respect to $\nu_{f_\E;T}$;
since this function is bounded by 1, we can again use the
dominated convergence theorem to say that the integral vanishes as
$N \rightarrow \infty$. One cannot give a more explicit bound for
this term without more information about the relationship between
$\mu$ and $\H$. In any case, we have convergence of the bias term
to 0.

Now, we can complete the proof of Theorem 1. For any $\eta>0$, we
must be able to show that $\lim_{N\rightarrow\infty}
P\{\|f_{Z_N,\lambda_N}-f_\E\|_\H \geq \eta\} = 0$. So, let us
choose an arbitrary fixed value $\eta$. The bias term vanishes as
$N \rightarrow \infty$ and $\lambda_N \rightarrow 0$, so there
must exist an $N_0$ so that for $N>N_0$, $\lambda_N$ is
sufficiently small, so that the bias term is bounded by $\eta/2$.
Thus, we consider the bias term bounded by $\eta/2$ in the limit
as $N \rightarrow \infty$; since this term does not depend on the
data, the bound clearly holds with probability 1. Now, we must
choose $\varepsilon_N$ so that the variance term is bounded by
$\eta/2$. Using the bound in (\ref{thm1variancebd}), we choose
$\varepsilon_N = \frac{\eta^2 \lambda_N}{8}$. The corresponding
probability $p_N$ is then given by ($\ref{thm1p}$),

\begin{eqnarray*}
p_N = \frac{64 M C^2 \kappa^2 + 8 M^2 \lambda_N}{N \lambda_N
\varepsilon_N^2} = \frac{64(64 M C^2 \kappa^2 + 8 M^2
\lambda_N)}{N \lambda_N^3  \eta^4}.
\end{eqnarray*}

We need $p_N$ to vanish as $N \rightarrow \infty$; this is
satisfied if $N \lambda_N^3 \rightarrow \infty$ as
$N\rightarrow\infty$. Also, there must exist an $N_0$ such that
for $N>N_0$, we have $N\geq \frac{8M^2}{\varepsilon_N^2}$; we need
this in order to use the Algorithmic Stability Theorem. Thus,
Theorem 1 is proved. \qed

\section{Proof of Theorem 2}

This section contains the proof of Theorem 2. First, some
notation. The positions $\mthx_1,..,\mthx_N$ will be considered
fixed throughout this section.\\
\noindent$ \text{\underline{def}}\;\;\; P:\H \longrightarrow
\R^N\\
\text{}\hspace{40pt} f \longmapsto
(f(\mthx_1),f(\mthx_2),..,f(\mthx_N)) $\\

\noindent The `evaluation operator' $P$ evaluates a function $f$
at each position $\mthx_i$ in the data set. Note that $P$ `loses
information' about a function $f$ by evaluating it at only $N$
points. That is, $Ker\;P$ is a nontrivial subspace of $\H$. The
adjoint $P^*:\R^N \longrightarrow \H $ of the operator $P$ is
given by $P^*:(c_1,..,c_N) \longmapsto \sumi c_i K_{\mthx_i}$.
One can show that $P$ is a bounded operator, with
$\|P\|_{\mathcal{L}(H,H)}=\|P^*P\|_{\mathcal{L}(\H,\H)}^{1/2}\leq
(\maxj \sumi |G_{ij}|)^{1/2}.$
The operator $P^*P$ is automatically positive and self-adjoint. We
will later use the spectral theorem on the bounded self-adjoint
operator $P^*P$.

We start the proof of Theorem 2 with the following Lemma.\\

\noindent\textbf{Lemma 2.1}: \textit{The following
characterizations of $\overline{f}$ are equivalent}:
\begin{enumerate}
    \item $\overline{f}$ satisfies:\\
    \textbf{(i)} $\overline{f}(\mthx_i)=\tilde{f}(\mthx_i)$ for
    $i=1,..,N$, and\\
    \textbf{(ii)}
    $ \|\overline{f}\|_\H \leq \|g\|_\H \;\;\forall g \in \H $ that satisfy $
    g(\mthx_i)=\tilde{f}(\mthx_i). $

    \item $\overline{f}$ satisfies:
    \begin{equation}\label{lemma2.1}
    \overline{f} = \sumi \tilde{f}(\mthx_i) W_{\mthx_i}, \text{ where
    }W_{\mthx_i} = \suml G^{-1}_{i\ell} K_{\mthx_\ell}
    \end{equation}

    \item $\overline{f}$ satisfies:\\
    \textbf{(i)} $\overline{f}(\mthx_i)=\tilde{f}(\mthx_i)$ for
    $i=1,..,N$, and\\
    \textbf{(iii)}
    $\forall h \in Ker \;P$ \; we have \; $(\overline{f},h)_\H=0$.
\end{enumerate}

\begin{proof} We will show 1. $\rightarrow$ 2. $\rightarrow$ 3.
$\rightarrow$ 1.\\
\noindent 1. $\leftrightarrow$ 2. First, we show that the function
described in 1. is unique. From the reproducing property, we know
that $\overline{f}$ has nonzero components along each of the
$K_{\mthx_i}$'s for which $\overline{f}(\mthx_i)\neq 0$. Since
$\H$ is a Hilbert space, we can always decompose $\overline{f}$
into a component $f_\|$ within the span of the $K_{\mthx_i}$'s and
a component $f_\perp$ orthogonal to each $K_{\mthx_i}$ (where $i
\in 1,..,N$). Now, $\|\overline{f}\|_\H^2 = \|f_\|\|_\H^2 +
\|f_\perp\|_\H^2 \geq \|f_\|\|_\H^2$. Thus, if $\|f_\perp\|_\H
\neq 0$, then $\overline{f}$ no longer has minimal norm and
contradicts property \textbf{(ii)}. The component of
$\overline{f}$ along each of the $K_{\mthx_i}$'s is determined by
the value of $\overline{f}(\mthx_i)$. So, functions $f$ that
satisfy both \textbf{(i)} and \textbf{(ii)} can be written $f =
f_\| = \sum_{i=1}^{N} \alpha_i K_{\mthx_i}$ for the fixed values
of $\alpha_i$, $i=1,..,N$. In particular, the $\alpha_i$'s must
satisfy:
\begin{eqnarray*}
\sumi \alpha_i G_{ij} = \sumi \alpha_i K_{\mthx_i}(\mthx_j) =
f(\mthx_j)\;.
\end{eqnarray*}
\noindent Thus, the function described in 1. is unique. It is
straightforward to see that the function described in 2. is
exactly the function described in 1. Evaluating the right side of
(\ref{lemma2.1}) at $\mthx_j$, we obtain
\begin{eqnarray*}
\overline{f}(\mthx_i) = \sumj \tilde{f}(\mthx_j) \suml G^{-1}_{j
\ell}G_{\ell i} = \tilde{f}(\mthx_i).
\end{eqnarray*}
Moreover, the function described in 2. lies entirely within the
span of the $K_{\mthx_i}$'s. Therefore it obeys \textbf{(i)} and
\textbf{(ii)} and we have 1. $\leftrightarrow$ 2.\\

\noindent 2. $\rightarrow$ 3. Because 2. $\rightarrow$ 1.,
\textbf{(i)} is
satisfied. We just need to show \textbf{(iii)}.\\
\noindent For any $h \in Ker \; P,$
\begin{eqnarray} \nonumber
h(\mthx_\ell)&=&0 \;\;\; \forall\; \ell \in 1,..,N
\\\nonumber (h,K_{\mthx_\ell})_\H&=&0 \;\;\;
\forall \; \ell \in 1,..,N \text{ by the reproducing
property,}
\\\nonumber (h,W_{\mthx_i})_\H&=&0 \;\;\; \forall \; \ell, i
\in 1,..,N \text{ because the }W_{\mthx_i}\text{'s are each a
linear combination of the }K_{\mthx_\ell}\text{'s}.
\\\nonumber (h,\overline{f})_\H&=&0 \text{ because } \overline{f}
\text{ is a linear combination of the } W_{\mthx_i}\text{'s, thus
\textbf{(iii)} holds}.
\end{eqnarray}

\noindent 3. $\rightarrow$ 1. Here, \textbf{(i)} is automatic, so
we need to check
\textbf{(ii)}. \\

\noindent$
\begin{array}{ll}
\text{Take arbitrary } g \in \H \text{ with: }& g(\mthx_i) =
\tilde{f}(\mthx_i) = \overline{f}(\mthx_i)
\text{ for } $i=1,..,N$. \\
\text{Then,}&\; g-\overline{f} \in Ker \; P.\\
\text{From assumption \textbf{(iii)},} &(\overline{f},g-\overline{f})_\H=0,\\
\text{and thus }&(\overline{f},g)_\H = \|\overline{f}\|_\H^2.\\
\end{array}
$

\vspace{5pt} \noindent \;\;Now,
\begin{eqnarray}\nonumber
\|g\|_H^2 &=& \|g-\overline{f}+\overline{f}\|_\H^2\\\nonumber &=&
\|g-\overline{f}\|_\H^2 + \|\overline{f}\|_\H^2 +
2(g-\overline{f},\overline{f})_\H\\\nonumber
&=&\|g-\overline{f}\|^2_\H + \|\overline{f}\|_\H^2 \\\nonumber
&\geq& \|\overline{f}\|_\H^2, \text{ \;\;with equality only if }
g=\overline{f}.
\end{eqnarray}

Thus we have $1. \rightarrow 2. \rightarrow 3. \rightarrow 1.$, so
Lemma 2.1 is proved.\qed
\end{proof}

Back to the proof of Theorem 2. The functional
$L_{Z_t,\lambda_t}(f)$ in the \textit{Main Algorithm} expressed in
terms of $P$ becomes:
\begin{eqnarray*}
 L_{Z_t,\lambda_t}(f)&=&\frac{1}{N}\Big\|P f - (P \tilde{f} + \frac{1}{t}\textbf{b})
 \Big\|^2_{\ell_2} + \lambda_t \|f\|_\H^2
\end{eqnarray*}

\noindent The minimizer of $L_{Z_t,\lambda_t}(f)$ must satisfy
$\frac{\partial}{\partial \gamma} L_{Z_t,\lambda_t}(f+\gamma h)
\big{|}_{\gamma=0} = 0$. In other words, the first variational
derivative of $L_{Z_t,\lambda_t}(f)$ is 0 at its minimizer.
Recalling that $P \tilde{f} = P
\overline{f}$, 
this minimization problem becomes:
\begin{eqnarray*}
0&=&\frac{\partial}{\partial \gamma} \big[ \frac{1}{N}(P f +\gamma
P h - P \overline{f} -\frac{1}{t}\mathbf{b},P f +\gamma P h - P
\overline{f} - \frac{1}{t}\mathbf{b})_{\ell_2} + \lambda_t \|f+
\gamma h\|^2_\H \big] \Big|_{\gamma=0}\\
&=&2\frac{1}{N}(P h, P f - P \overline{f}
-\frac{1}{t}\mathbf{b})_{\ell_2} + 2\lambda_t(f,h)_\H \\&=&
2\Big(h,\frac{1}{N}P^*P f - \frac{1}{N}P^* P \overline{f} -
\frac{1}{N}P^*\Big(\frac{1}{t}\mathbf{b}\Big) + \lambda_t
f\Big)_\H .
\end{eqnarray*}
This must be true for any function $h$, so $\frac{1}{N}P^* P f -
\frac{1}{N}P^* P \overline{f} -\frac{1}{N}P^*
\Big(\frac{1}{t}\mathbf{b}\Big) + \lambda_t f = 0$, implying
\begin{eqnarray*}
f = \overline{f}-(\frac{1}{N}P^* P + \lambda_t)^{-1}\lambda_t
\overline{f}+ (\frac{1}{N}P^* P + \lambda_t)^{-1}\frac{1}{N}P^*
\Big(\frac{1}{t}\mathbf{b}\Big) \;.
\end{eqnarray*}
It follows that
\begin{eqnarray} \nonumber
 \|f-\overline{f}\|_\H
&\leq& \|(\frac{1}{N}P^* P +\lambda_t)^{-1}\lambda_t
\overline{f}\|_\H + \frac{1}{Nt}\|(\frac{1}{N}P^* P +
\lambda_t)^{-1}P^*\|_{\mathcal{L}(\ell_2,\H)}
\|\mathbf{b}\|_{\ell_2}.\;\;\;\label{thm2terms}
\end{eqnarray}

In order to show stability, we bound the two terms on the right of
(\ref{thm2terms}), and construct these bounds so they vanish as
$t\rightarrow\infty$. That is, we need to bound the norms above.
To accomplish this, we will use the spectral theorem on the
bounded self-adjoint operator $P^* P$.

To bound the first term in equation (\ref{thm2terms}), recall that
the operator obtained from the function
$\phi_t(z)=(\frac{1}{N}z+\lambda_t)^{-2}\lambda_t^2$ of the
self-adjoint operator $P^*P$ is self-adjoint. Also, since $P^*P$
is a positive operator, the spectrum $spec(P^*P)$ of the operator
$P^*P$ is concentrated on $\R_+ \cup \{0\}$. Using the spectral
measure $\nu_{\overline{f};P^*P}(z)$ on the spectrum $spec(P^*P)$,
we find:
\begin{eqnarray*}
\|(\frac{1}{N}P^* P +\lambda_t)^{-1} \lambda_t \overline{f}\|^2_\H
&=& ([(\frac{1}{N}P^* P +\lambda_t)^{-1} \lambda_t]^2
\overline{f}, \overline{f})_\H.
\\&=& \int_{spec(P^* P)} \left(\frac{\lambda_t}{\frac{1}{N}z +
\lambda_t}\right)^2 d \nu_{\overline{f};P^*P}(z)
\end{eqnarray*}
Because $\lambda_t \limt 0$, we have $\phi_t(z) =
(\frac{1}{N}z+\lambda_t)^{-2}\lambda_t^2 \limt 0$ for all $z \in
\R_+ \setminus \{0\}$. By Lemma 2.1, we know that $\overline{f}
\perp Ker\; P$, and since $Ker\; P = Ker \;P^* P$, we know that
$\nu_{\overline{f};P^*P} (\{0\}) =0$. Since $\phi_t(z) \leq 1$ for
all $z\in spec(P^*P) \subset \R_+$, it then follows from the
dominated convergence theorem that $\|(\frac{1}{N}P^* P
+\lambda_t)^{-1} \lambda_t \overline{f}\|^2_\H \limt 0$. One
cannot give a more explicit bound for this first term; it would
require more specific knowledge of the relationship between $\mu$
and $\H$. In any case, we have achieved our goal in showing that
the first term of (\ref{thm2terms}) vanishes as $t \rightarrow
\infty$.

We need the second term in equation (\ref{thm2terms}) to vanish
also. Recall that for operator $Q: H_1 \rightarrow H_2$, it is
true that
$\|Q^*\|_{\mathcal{L}(H_2,H_1)}=\|Q^*Q\|^{1/2}_{\mathcal{L}(H_1,H_1)}$,
and that a continuous real function of a self adjoint operator
such as $P^*P$ is self adjoint.
\begin{eqnarray*}
\|(\frac{1}{N}P^* P +
\lambda_t)^{-1}P^*\|_{\mathcal{L}(\ell_2,\H)} = \|(\frac{1}{N}P^*
P +\lambda_t)^{-1} P^* P (\frac{1}{N}P^* P +
\lambda_t)^{-1}\|^{1/2}_{\mathcal{L}(\H,\H)}.
\end{eqnarray*}
\noindent We use the spectral theorem for the bounded self-adjoint
operator $A=P^* P$, namely the fact
$\|\phi(A)\|_{\mathcal{L}(H,H)} \leq sup \{ |\phi(z)| ; z \in
spec(A)\}$ where $\phi(z) = \frac{z}{(\frac{1}{N}z+\lambda_t)^2}$
here. The maximum value of $\phi(z)$ occurs at $z=N \lambda_t$,
and it is $\frac{N}{4 \lambda_t}$. Thus,
\begin{eqnarray*}
\|(\frac{1}{N}P^* P +
\lambda_t)^{-1}P^*\|_{\mathcal{L}(\ell_2,\H)} &\leq&
\frac{\sqrt{N}}{2\sqrt{\lambda_t}}\\
\frac{1}{Nt}\|(\frac{1}{N}P^* P +
\lambda_t)^{-1}P^*\|_{\mathcal{L}(\ell_2,\H)}
\|\textbf{b}\|_{\ell_2}&\leq&
\frac{\sqrt{N}}{2Nt\sqrt{\lambda_t}}\|\textbf{b}\|_{\ell_2} \leq
\frac{1}{2t\sqrt\lambda_t} b^{max} \text{\;a.s.}
\end{eqnarray*}
As long as we design $\lambda_t$ so that $t \sqrt{\lambda_t} \limt
\infty$, then we have the desired convergence of this term to 0.
We are done with the second term of equation (\ref{thm2terms}).
Theorem 2 is proven.\qed

\section{Conclusion}
We have proved stability for the regularized least squares
regression algorithm, for the sense in which inverse problems are
examined.
We have shown stability for this algorithm in two cases: the case
when the number of data points $N$ is a constant, and the case
where $N \rightarrow \infty$. It is important that our algorithm
is stable in this sense, because we do not want any inherent error
in the algorithm's output. Neither a small amount of noise in the
data nor a small amount of regularization should drastically
influence the algorithm's output.



We hope that the reader will gain more from our result than the
knowledge that regularized least squares regression is stable in
the inverse operator sense. We have found the particular methods
introduced in the proofs of Theorem 1 and Theorem 2 useful for
various learning problems, especially those which require the
convexity of learning functionals or convergence of learning
algorithms. Namely, we demonstrate two methods for showing that
the minimizers of two learning functionals are close: use of the
spectral theorem, and the technique of Lemma 1.1, which can both
be generally applied to other learning algorithms.

\section{Acknowledgements} The author would like to express
infinite gratitude to Ingrid Daubechies.


\end{document}